\documentclass{article}

\usepackage{xcolor}
\usepackage{microtype}
\usepackage{graphicx}
\usepackage{subfigure}
\usepackage{booktabs} 

\usepackage{layouts}

\usepackage{hyperref}

\usepackage[accepted]{icml2021}

\icmltitlerunning{A study on the plasticity of neural networks}

\usepackage[font=small,skip=3pt]{caption}

\begin{document}
\twocolumn[
\icmltitle{A study on the plasticity of neural networks}

\icmlsetsymbol{equal}{*}

\begin{icmlauthorlist}
\icmlauthor{Tudor Berariu}{icl}
\icmlauthor{Wojciech Czarnecki}{dm}
\icmlauthor{Soham De}{dm}
\icmlauthor{Jorg Bornschein}{dm}
\icmlauthor{Samuel Smith}{dm}
\icmlauthor{Razvan Pascanu}{dm}
\icmlauthor{Claudia Clopath}{icl,dm}
\end{icmlauthorlist}
\icmlaffiliation{icl}{Imperial College London, Department of Bioengineering, London, UK}
\icmlaffiliation{dm}{DeepMind, London, UK}
\icmlcorrespondingauthor{Tudor Berariu}{t.berariu19@imperial.ac.uk}
\icmlkeywords{Machine Learning, ICML, Continual Learning, Generalisation}

\icmlkeywords{Machine Learning, ICML}

\vskip 0.3in
]

\printAffiliationsAndNotice{} 

\begin{abstract}
One aim shared by multiple settings, such as continual learning or transfer learning, is to leverage previously acquired knowledge to converge faster on the current task. Usually this is done through fine-tuning, where an implicit assumption is that the network maintains its \textit{plasticity}, meaning that the performance it can reach on any given task is not affected negatively by previously seen tasks. It has been observed recently that a pretrained model on data from the same distribution as the one it is fine-tuned on might not reach the same generalisation as a freshly initialised one. We build and extend this observation, providing a hypothesis for the mechanics behind it. We discuss the implication of losing plasticity for continual learning which heavily relies on optimising pretrained models.
\end{abstract}

\section{Introduction}
\label{sec:intro}

Continual learning is concerned with training on non-stationary data. In a practical description, an agent learns a sequence of tasks, being restricted to interact with only one at a time. There are several desiderata for a successful continual learning algorithm. First, agents should not forget previously acquired knowledge, unless capacity becomes an issue or contradicting facts arrive. Second, such an algorithm should be able to exploit structural similarity between tasks and exhibit accelerated learning. Third, backward transfer should be possible whenever new knowledge helps generalisation on previously learnt tasks. Fourth, successful continual learning relies on an enduring capacity to acquire new knowledge, therefore learning now should not impede performance on future tasks.

\begin{figure}
    \centering
    \includegraphics[width=\linewidth]{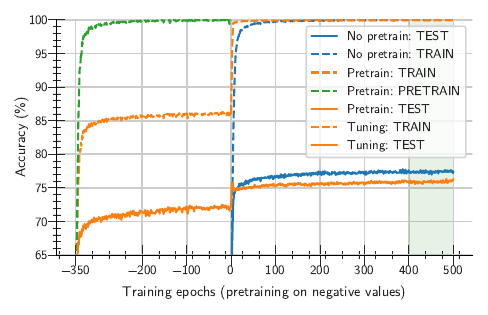}
    \caption{Our reproduction of the core experiment performed by \cite{ash2019difficulty}. A ResNet-18 model is pretrained on half of the CIFAR 10 training data, and then tuned on the full training set. It generalises worse than the model trained from scratch.}
    \label{fig:demo}
\end{figure}

In this work we focus on \textit{plasticity}, namely the ability of the model to keep \textit{learning}. There are different nuances of \textit{not being able to learn}. A neural network might lose the capacity to minimise the training loss for a new task. For example, PackNet~\citep{PackNet} eventually gets to a point where all neurons are frozen and learning is not possible anymore. In the same fashion, accumulating constraints in EWC~\citep{EWC} might lead to a strongly regularised objective that does not allow for the new task's loss to be minimised. Alternatively, learning might become less data efficient, referred to as \textit{negative forward transfer}, an effect often noticed for regularisation based continual learning approaches. In such a situation one might still be able to reduce training error to $0$ and obtain full performance on the new task, is just learning is considerably slower. Lastly, a third meaning and the one we are concerned with, is that while training error can be reduced to zero, and irrespective to how fast the model learns, the optimisation might lead to a poor minimum which achieves lower generalisation performance. \footnote{\emph{Oct 2023} -- The authors acknowledge, and this paragraph was trying to highlight that \emph{loss of plasticity} as a term has been used to describe two different phenomena. On one hand we have the phenomenon studied in this work, where the issue is not the inability of the neural network to reduce error on training set, but rathter inability to generalize. I.e. the warm-started model will give you worse generalization even if the training error can be reduced equally well as the freshly initialized model. The other form of the problem (e.g. studied in \cite{dahore21}) talks about inability of the system to optimize. While the underlying causes could be similar, this is unlikely, and until shown to be the same problem we acknowledge that they should be treated as separate problems (as per the conclusion the authors had with Aaron Courville). In terms of the name, it is not clear what the correct answer is. Partially the problem is that the concept of plasticity comes from neuroscience, and attributed to inability to learn in machine learning, which is an ambigous term. Learning as a concept relies on optimization, but requires ability to generalize too.  While we do not know how to resolve the name conflict, we want to make the reader aware of it.    }
We define as the \textit{generalisation gap} the difference in performance a pretrained model can obtain --- e.g. one that had learnt a few tasks already --- versus a freshly initialised one, without constraining the number of updates. Note that this is similar to the notion of intransigence proposed by \cite{Chaudhry2018RiemannianWF}, but instead making the comparison against a model trained only on the new data, rather than a multi-task solution. Our focus is on understanding if a \textit{generalisation gap} exists, whether it is positive or negative. 
The \textit{transfer learning} dogma used to indicate a positive gap, arguing that pretraining on large sets of data provides a good initialisation for a related target task.
Recently, \cite{he2019rethinking} showed that this does not necessarily hold,
reporting state-of-the-art results with randomly initialised models, although 
at worse sample complexity.
\cite{ash2019difficulty} considered an extreme transfer scenario, where an agent is pretrained on data from the same distribution as the target task, and reported a negative generalisation gap. See Figure~\ref{fig:demo} for our reproduction of this finding. We build on this result in this work, trying to further expand the empirical evidence on which factor affect the generalisation gap and take a first step towards understanding its root causes.

\citeauthor{igl2021transient} extend the observation that data non-stationarity affects asymptotic generalisation performance to reinforcement learning scenarios. Although both \citeauthor{ash2019difficulty}, and \citeauthor{igl2021transient} propose solutions to close the generalisation gap, the reasons for its occurrence in the first place are still unclear.
We argue that it can have a considerable impact on how we approach continual learning, and one should track to what extend it affects the algorithms we have.

\section{Generalisation gap - Experiments}
\label{sec:results}

In this section we present our analysis of the generalisation gap, and detail a series of experiments we argue are indicative of the aggravation of the phenomenon in continual learning. We ask a couple of questions and provide empirical evidence to answer them, such as: How much pretraining is too much? Is this negative effect additive when pretraining consists of several stages? Is there a way to leverage the pretrained parameters for faster tuning? 

We start with the same setup as in \cite{ash2019difficulty} training deep residual networks \cite{He_2016_CVPR} to classify the CIFAR 10 data set. We use the average test accuracy in the last 100 training epochs (see the green box in Figure~\ref{fig:demo}) to compare different setups. We mention below the relevant details for each experiment, and we offer a full description of the empirical setup in Appendix~\ref{sec:cifar}.

\begin{figure}[h!tb]
    \centering%
    \includegraphics[width=\linewidth]{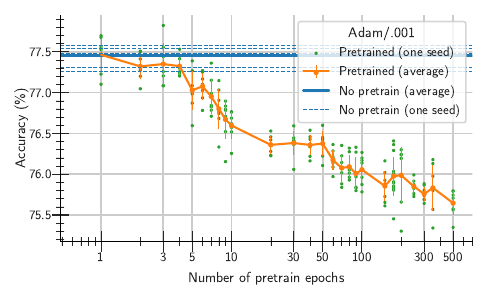}
    \caption{Average test accuracy in the last 100 epochs of tuning after pretraining the model for different numbers of epochs using Adam with a constant learning rate ($10^{-3}$) for both phases. }
    \label{fig:adam}
\end{figure}

\paragraph{Does the optimization algorithm affect the gap?} Different optimizers have particular advantages in escaping sub-optimal regions. We reproduced the warm start experiment for a couple of different optimisers using constant learning rates: Adam, RMSprop, SGD, and SGD with momentum (Figures~\ref{fig:adam},~\ref{fig:rmsprop},~\ref{fig:sgd},~and~\ref{fig:msgd}). \citeauthor{ash2019difficulty} reported similar results. The fact that the generalisation gap manifests in all cases supports the observation that it is rather a problem with the quality of the local minima 
than one of finding appropriate descending trajectories.

\paragraph{How many pretraining steps are needed to produce a generalisation gap?} We investigated how much does the gap depend on a large number of optimisation steps in the pretraining stage. Would early stopping close the gap? In our experiments, although a larger number of pretraining steps hurt generalisation more in the tuning stage, just a few passes through the data are enough to observe a gap (5-10 for Adam, which is even before reaching 100\% accuracy). In conclusion the gap is there before any reason to do early stopping. This is consistent across all considered optimisers (Figures~\ref{fig:adam},~\ref{fig:rmsprop},~\ref{fig:sgd},~and~\ref{fig:msgd}). A similar experiment was reported in \cite{ash2019difficulty}.

\begin{figure}[h!tb]
    \centering%
    \includegraphics[width=\linewidth]{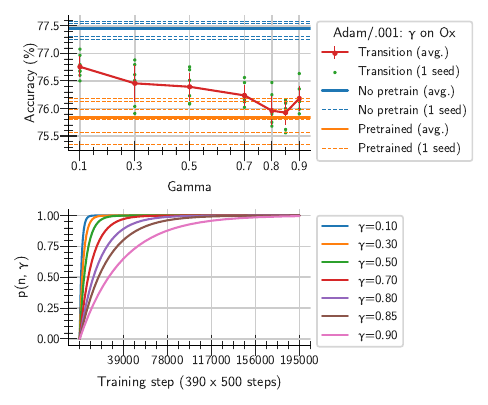}
    \caption{Models trained in a single stage where each example is individually sampled with probability $p=1-\gamma^{50 n/N}$ from the full training data, and with probability $1-p$ from the pretrain set ($n$ is the current step, while $N$ represents the total number of steps -- the equivalent of 500 epochs). A few more details in Section~\ref{sec:blending}.}
    \label{fig:blending}
\end{figure}

\paragraph{Is the gap still there when data distribution slides smoothly?}
We tested whether a smooth transition from the pretraining subset to the full training set would remove the generalisation gap. But, as Figure~\ref{fig:blending} shows, the generalisation gap manifests even for a small number of epochs with biased sampling. This might have profound implications in reinforcement learning where the data distribution changes slowly during training, as the policy collecting the data changes.

Concluding from the last two sets of experiments, a transient bias in the data distribution significantly impacts generalisation performance.

\begin{figure}[h!tb]
    \centering%
    \includegraphics[width=\linewidth]{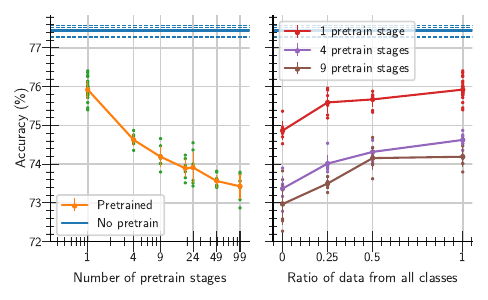}
    \caption{The same model (ResNet18) was trained in multiple stages. All but the last pretrain stages consisted of a number of steps proportional with the number of examples and sufficient to reach 100\% accuracy on train. Right: New data for a particular stage has a ratio of examples drawn uniformly from the training set, and the rest from classes designated for that stage (see Section~\ref{sec:splits} for details).}
    \label{fig:stages}
\end{figure}

\paragraph{Do multiple pretrain stages and/or class ordering matter?} Continual learning is concerned with possibly unlimited changes in the data distribution. It is natural to ask whether the loss in generalisation performance observed as a consequence of a single pretraining stage is aggravated when data is incrementally added in more steps. In order to answer this question we divided the data set in multiple splits training the model in stages. We show in Figure~\ref{fig:stages} that the final generalisation performance (the test accuracy achieved in the last stage training on the full training set) degrades with the number of splits.

To get even closer to the usual continual learning setup, we considered splits of the training set having some level of class imbalance, therefore exhibiting larger differences between the data distributions considered at consecutive stages (Figure~\ref{fig:stages}, right). We tested for splits ranging from class partitions where each stage would bring data from one or more new classes to the uniform subsampling from the training set considered so far (see Section~\ref{sec:splits} for the detailed methodology used to split the data set). We noticed that higher discrepancies between training stages lead to worse generalisation.

We argue that this observations point out a core difficulty of continual learning. When saving data from the past is feasible, retraining models seems a better strategy than using pretrained models.

\begin{figure}[h!tb]
    \centering%
    \includegraphics[width=\linewidth]{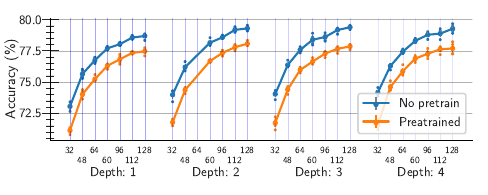}
    \caption{Average performance on the test set for residual networks of various depths and widths. See Section~\ref{sec:resnets} for details on the models' architectures.}
    \label{fig:architectures}
\end{figure}

\paragraph{How do model width and/or the depth change the generalisation gap?}
We investigated whether increasing the capacity of the model helps recovering the generalisation performance of a randomly initialised model. We show in Figure~\ref{fig:architectures} that even for very deep and wide models there is a significant gap between pretrained models and those trained from random initialisations.
\begin{figure}[h!tb]
    \centering%
    \includegraphics[width=\linewidth]{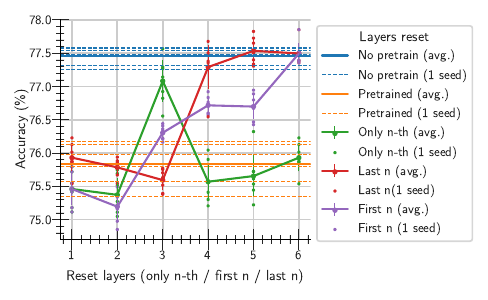}
    \caption{Performance of models for which a subset of layers were reset after pretraining. 1 represents the first convolution, 2-5 are the four modules, and 6 is the fully connected output layer.}
    \label{fig:reset}
\end{figure}
\paragraph{Which pretrained parameters should be kept for tuning not to have a gap?} Knowing that tuning the whole model leads to poor generalisation performance we ask what the best strategy is for taking advantage from the pretrained model?
We conducted a series of experiments in which we re-sample the parameters of some layers from the same distribution used at initialisation. In our tests with ResNet-18 models on CIFAR 10, resetting just a small subset of the layers is not enough to fully recover the gap. Our experiments, summarised in Figure~\ref{fig:reset}, indicate that in order to close the gap the top part of the model must be reinitialised.
Therefore it might be advantageous to keep the first $k$ layers (as it is usually done in transfer learning), but in our experiments, $k$ is quite small. Moreover, it seems that there is no advantage in terms of training speed to keep the pretrained layers (see Section~\ref{sec:reset_details} for details.)
\section{A possible account for the generalisation gap: Two Phases of Learning}
\label{sec:two_phases}

One plausible hypothesis for the occurrence of the generalisation gap stems from the flat versus sharp minima view on generalisation \cite{hochreiter1997flat}. Precisely, local minima which exhibit low curvature, and wide basins of attraction generalise better than sharp ones. This could be motivated from an information theoretic perspective: flat minima require less precision to be described (the minimum description length argument made by \cite{hochreiter1997flat}); or by thinking about the stability around that point: flat minima are affected less by perturbations in the inputs. Although there is no formal definition for flatness, previous works proposed quantities such as the largest eigenvalue of the Hessian \cite{keskar2017on}, or the local entropy \cite{chaudhari2019entropy} to gauge it. 

If optimisation were to follow the gradient flow (the infinitesimally precise path determined by the gradient), then the minimum it converges to would be determined by the initial random initialisation. In practice optimisation diverges from that path. This is due to the noise induced by the randomness in the mini-batch approximation of the loss function, and by the amplitude of the update step. As a consequence the training dynamics allegedly traverse two phases.
In the early \textit{exploration} phase, the parameters ``bounce'' form the vicinity of a critical point to another until they land in the basin of attraction of a minimum that is wide enough to trap the optimisation.

Once the parameters get stuck in the basin of attraction of some minumum, training goes into a \textit{refinement} phase,  where parameters converge to the said critical point. In this phase optimisation follows the gradient flow. 

Several works bring supporting evidence for the \emph{two phases of learning} hypothesis. \cite{achille2018critical} also identifies two phases by analysing the information stored in the weights. Their observations are consistent with the sharp versus flat minima view as the the Fisher Information Matrix used to measure connectivity is also indicative about curvature. \cite{golatkar2019time} reveal that regularisation has an impact only in the initial phase, while \cite{gur2018gradient} show that after an early regime gradients reside in a small subspace that remains constant during training. More relevant works are mention in Section~\ref{sec:twophase_literature}.

Building on this hypothesis, it is natural to ask whether the generalisation gap can be explained by how pretraining affects the \textit{exploration} phase of learning. 

Note that the amount of exploration that learning is able to do in this initial stage is proportional to several factors, among which the more important ones are the learning rate and the inherent noise in the updates. The role of the learning rate is self-explanatory, it can be seen as scaling the amount of noise. 
As for the noise itself, there are multiple sources: the inherent noise in the data (labelling imprecisions, noise in the observations, irrelevant features), noise induced by the optimiser of choice (e.g. SGD introduces noise by relying on mini-batches), noise induced by the data augmentation procedure, etc.
The noise profile of gradients and parameters updates is important, and known to be non-Gaussian \cite{simsekli2019tail}, hence it is harder to replicate in practice.

We make the conjecture that, \emph{given that everything else stays the same, the pretraining stage can considerably reduce the amount of noise in the gradients during the tuning stage, which leads to weaker exploration and convergence to narrower minima, inducing the generalization gap.}

In particular, in the case of discriminative learning which has been the focus of this work, estimators tend to quickly become robust to many directions of variation in the data which are not relevant for the classification task. For example, the model starts ignoring non-discriminative features such as background patterns early in training. In fact this property is of great importance to the recent success of neural networks. Many architectural advances are more efficient in doing so leading to more robust models with better generalisation properties (e.g. the translation-invariant convolutions compared with fully-connected linear layers). Data augmentation plays a similar role.

However, these irrelevant features do potentially play a role in the initial stage of learning as a source of noise, forcing the optimisation process to focus on wider minima. When the model is pretrained, it becomes insensitive to some of these irrelevant features (or some easy to discern sources of noise). While it is true that when moving to the tuning stage the problem changes --- and hence the loss surface is different and there is likely no relationship between the two loss surfaces and their critical points (e.g. this is the case in a continual learning setting) --- the model will still be insensitive to some direction of change (either in the input space or in the latent space). Even if those direction of variations are not relevant for the new task, it does mean that in the early stage of tuning there will be considerably less noise, and hence potentially less exploration. This means that optimisation in the tuning stage will converge to a narrower minimum, which will generalise less, leading to the generalization gap. 

\begin{figure}[h!tb]
    \centering%
    \includegraphics[width=\linewidth]{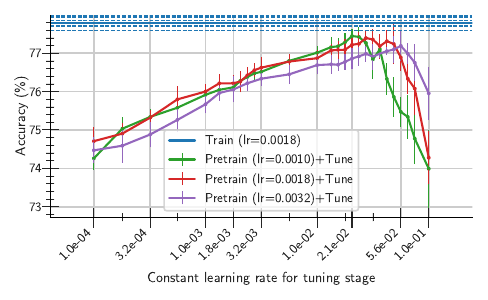}
    \caption{In this figure we show the performance reached by tuned models for three specific constant learning rates used during pretraining (the $0.0018$ is the one that generalizes best). Each point is the average of 9 seeds. The horizontal lines represent the performance achieved by randomly initialised models. Therefore each distance between a circle and the horizontal lines represents the gap for a particular pair of learning rates.}
    \label{fig:lr_grid}
\end{figure}

To test this hypothesis we check whether increasing the learning rate -- which would magnify the remaining  noise in the parameter updates-- during the tuning stage helps.  Figure~\ref{fig:lr_grid} shows that a 10x larger learning rate reduces the gap substantially. The rest of the performance gap could be explained by the fact that a high constant learning rate does not benefit from the \emph{refinement} phase. While further empirical evidence is required to validate this hypothesis, this result is encouraging. Under the assumption that this is the cause for the gap, one question to be asked is why does the pretraining phase reduce gradient noise for the tuning phase. The answer might rest in two observations: 1) the strong data overlap between the two stages, and 2) neural networks tend to filter early on information in the input that is not discriminative. If the non-discriminative dimensions  are already filtered out by pretraining, the tuning stage might not benefit from the noise they would induce.

\section{Conclusions}
\label{sec:future}

We build on~\cite{ash2019difficulty} and study the generalisation gap induced by pretraining the model on the same data distribution. We extend the original results, by looking at robustness of this gap to smooth transition between data distributions, multiple stages of pretraining, model size or resetting parts of the pretrained model. We take a first step towards understanding this phenomenon by asking whether it is related to the two phases of learning hypothesis. 

The existence of this generalisation gap suggest that continual learning might be hurt by using compact models that get finetuned on multiple tasks. We argue that tracking the generalisation gap represents a new facet of forward transfer that has not generally been measured or tracked in the literature.

\bibliography{biblio}

\begin{thebibliography}{18}
\providecommand{\natexlab}[1]{#1}
\providecommand{\url}[1]{\texttt{#1}}
\expandafter\ifx\csname urlstyle\endcsname\relax
  \providecommand{\doi}[1]{doi: #1}\else
  \providecommand{\doi}{doi: \begingroup \urlstyle{rm}\Url}\fi

\bibitem[Achille et~al.(2019)Achille, Rovere, and Soatto]{achille2018critical}
Achille, A., Rovere, M., and Soatto, S.
\newblock Critical learning periods in deep networks.
\newblock In \emph{International Conference on Learning Representations}, 2019.
\newblock URL \url{https://openreview.net/forum?id=BkeStsCcKQ}.

\bibitem[Ash \& Adams(2019)Ash and Adams]{ash2019difficulty}
Ash, J.~T. and Adams, R.~P.
\newblock On the difficulty of warm-starting neural network training.
\newblock \emph{arXiv preprint arXiv:1910.08475}, 2019.

\bibitem[Chaudhari et~al.(2019)Chaudhari, Choromanska, Soatto, LeCun, Baldassi,
  Borgs, Chayes, Sagun, and Zecchina]{chaudhari2019entropy}
Chaudhari, P., Choromanska, A., Soatto, S., LeCun, Y., Baldassi, C., Borgs, C.,
  Chayes, J., Sagun, L., and Zecchina, R.
\newblock Entropy-sgd: Biasing gradient descent into wide valleys.
\newblock \emph{Journal of Statistical Mechanics: Theory and Experiment},
  2019\penalty0 (12):\penalty0 124018, 2019.

\bibitem[Chaudhry et~al.(2018)Chaudhry, Dokania, Ajanthan, and
  Torr]{Chaudhry2018RiemannianWF}
Chaudhry, A., Dokania, P.~K., Ajanthan, T., and Torr, P. H.~S.
\newblock Riemannian walk for incremental learning: Understanding forgetting
  and intransigence.
\newblock In \emph{ECCV}, 2018.

\bibitem[Dohare et~al.(2021)Dohare, Sutton, and Mahmood]{dahore21}
Dohare, S., Sutton, R., and Mahmood, A.~R.
\newblock Continual backprop: Stochastic gradient descent with persistent
  randomness.
\newblock \emph{arXiv preprint arXiv:2108.06325}, 2021.

\bibitem[Ghorbani et~al.(2019)Ghorbani, Krishnan, and
  Xiao]{ghorbani2019investigation}
Ghorbani, B., Krishnan, S., and Xiao, Y.
\newblock An investigation into neural net optimization via hessian eigenvalue
  density.
\newblock \emph{arXiv preprint arXiv:1901.10159}, 2019.

\bibitem[Golatkar et~al.(2019)Golatkar, Achille, and Soatto]{golatkar2019time}
Golatkar, A.~S., Achille, A., and Soatto, S.
\newblock Time matters in regularizing deep networks: Weight decay and data
  augmentation affect early learning dynamics, matter little near convergence.
\newblock In \emph{Advances in Neural Information Processing Systems}, pp.\
  10677--10687, 2019.

\bibitem[Gur-Ari et~al.(2018)Gur-Ari, Roberts, and Dyer]{gur2018gradient}
Gur-Ari, G., Roberts, D.~A., and Dyer, E.
\newblock Gradient descent happens in a tiny subspace.
\newblock \emph{arXiv preprint arXiv:1812.04754}, 2018.

\bibitem[He et~al.(2016)He, Zhang, Ren, and Sun]{He_2016_CVPR}
He, K., Zhang, X., Ren, S., and Sun, J.
\newblock Deep residual learning for image recognition.
\newblock In \emph{The IEEE Conference on Computer Vision and Pattern
  Recognition (CVPR)}, June 2016.

\bibitem[He et~al.(2019)He, Girshick, and Doll{\'a}r]{he2019rethinking}
He, K., Girshick, R., and Doll{\'a}r, P.
\newblock Rethinking imagenet pre-training.
\newblock In \emph{Proceedings of the IEEE International Conference on Computer
  Vision}, pp.\  4918--4927, 2019.

\bibitem[Hochreiter \& Schmidhuber(1997)Hochreiter and
  Schmidhuber]{hochreiter1997flat}
Hochreiter, S. and Schmidhuber, J.
\newblock Flat minima.
\newblock \emph{Neural Computation}, 9\penalty0 (1):\penalty0 1--42, 1997.

\bibitem[Igl et~al.(2021)Igl, Farquhar, Luketina, B{\"o}hmer, and
  Whiteson]{igl2021transient}
Igl, M., Farquhar, G., Luketina, J., B{\"o}hmer, W., and Whiteson, S.
\newblock Transient non- stationarity and generalisation in deep reinforcement
  learning.
\newblock In \emph{Proceedings of the International Conference on Learning
  Representations}. OpenReview, 2021.

\bibitem[Jastrzebski et~al.(2020)Jastrzebski, Szymczak, Fort, Arpit, Tabor,
  Cho, and Geras]{jastrzebski2020break}
Jastrzebski, S., Szymczak, M., Fort, S., Arpit, D., Tabor, J., Cho, K., and
  Geras, K.
\newblock The break-even point on optimization trajectories of deep neural
  networks.
\newblock \emph{arXiv preprint arXiv:2002.09572}, 2020.

\bibitem[{Keskar} et~al.(2017){Keskar}, {Mudigere}, {Nocedal}, {Smelyanskiy},
  and {Tang}]{keskar2017on}
{Keskar}, N.~S., {Mudigere}, D., {Nocedal}, J., {Smelyanskiy}, M., and {Tang},
  P. T.~P.
\newblock On large-batch training for deep learning: Generalization gap and
  sharp minima.
\newblock In \emph{ICLR 2017 : International Conference on Learning
  Representations 2017}, 2017.

\bibitem[Kirkpatrick et~al.(2017)Kirkpatrick, Pascanu, Rabinowitz, Veness, and
  et. al.]{EWC}
Kirkpatrick, J.~N., Pascanu, R., Rabinowitz, N.~C., Veness, J., and et. al.
\newblock Overcoming catastrophic forgetting in neural networks.
\newblock \emph{Proceedings of the National Academy of Sciences of the United
  States of America}, 114 13:\penalty0 3521--3526, 2017.

\bibitem[Li et~al.(2019)Li, Wei, and Ma]{li2019towards}
Li, Y., Wei, C., and Ma, T.
\newblock Towards explaining the regularization effect of initial large
  learning rate in training neural networks.
\newblock In \emph{Advances in Neural Information Processing Systems}, pp.\
  11669--11680, 2019.

\bibitem[Mallya \& Lazebnik(2017)Mallya and Lazebnik]{PackNet}
Mallya, A. and Lazebnik, S.
\newblock Packnet: Adding multiple tasks to a single network by iterative
  pruning.
\newblock \emph{2018 IEEE/CVF Conference on Computer Vision and Pattern
  Recognition}, pp.\  7765--7773, 2017.

\bibitem[Simsekli et~al.(2019)Simsekli, Sagun, and
  Gurbuzbalaban]{simsekli2019tail}
Simsekli, U., Sagun, L., and Gurbuzbalaban, M.
\newblock A tail-index analysis of stochastic gradient noise in deep neural
  networks.
\newblock In \emph{International Conference on Machine Learning}, pp.\
  5827--5837. PMLR, 2019.

\end{thebibliography}
\bibliographystyle{icml2021}
\newpage\newpage  ;

\newpage\pagebreak
\appendix
\section{Experiments on CIFAR-10}
\label{sec:cifar}

We detail below the experimental details for the observations made in Section~\ref{sec:results}. If it's not explicitly mention otherwise, a ResNet-18 model was trained on CIFAR 10 with batches of 128 examples, using an Adam optimiser with a constant learning rate of 0.001. The optimiser's statistics were reset between the warm up and the tuning phase. In the first stage (during warm up) the model was trained for 350 epochs on half of data. In the second stage the model is trained for 500 epochs on all training data. We report here the average test performance over all seeds during the last 100 training epochs. In all plots the error bars measure standard deviation.

Due to space constraints we don't show learning curves, but if it's not otherwise specified it's implied that accuracy is 100\% on the data used in training for both stages as in Figure~\ref{fig:demo}.

\subsection{Different optimisers}

\begin{figure}[h!tb]
    \centering%
    \includegraphics[width=\linewidth]{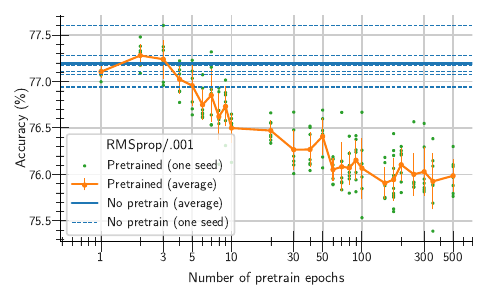}
    \caption{Final performance after warming up the model for different numbers of epochs using RMSProp with a constant learning rate for both phases.}
    \label{fig:rmsprop}
\end{figure}

\begin{figure}[h!tb]
    \centering%
    \includegraphics[width=\linewidth]{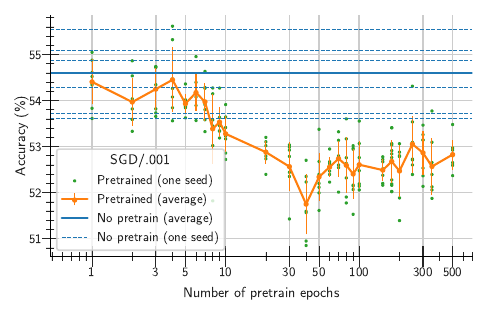}
    \caption{Final performance after warming up the model for different numbers of epochs using SGD with a constant learning rate for both phases.}
    \label{fig:sgd}
\end{figure}

\begin{figure}[h!tb]
    \centering%
    \includegraphics[width=\linewidth]{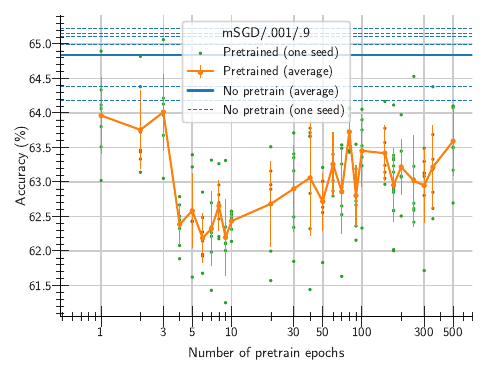}
    \caption{Final performance after warming up the model for different numbers of epochs using SGD with a constant learning rate and momentum for both phases.}
    \label{fig:msgd}
\end{figure}

In addition to Figure~\ref{fig:adam} in Section~\ref{sec:results} we show here how generalisation performance varies with the number of pretraining epochs on half of data for three additional optimisers: RMSprop in Figure~\ref{fig:rmsprop}, Stochastic Gradient Descent (SGD) in Figure~\ref{fig:sgd}, and SGD with a constant momentum (0.9) in Figure~\ref{fig:msgd}.

\subsection{Smooth transition between distributions.}
\label{sec:blending}

In the experiments presented in Figure~\ref{fig:blending} we trained models in a single stage of 500 epochs.
In this case we called an epoch a sequence of 390 update steps (which is the equivalent of 1 pass through the training data with a batch size of 128). Note that such an epoch is not a permutation of the data. Each example from each batch is individually sampled with probability $p$ from the full training set, and with probability $1-p$ from the pretrainig set.

\subsection{Class imbalance in the multiple stage setup}
\label{sec:splits}

Given (i) a training set $\mathcal{D}$ with examples from a set of classes $\mathcal{C}$, (ii) a number of pretraining stages $n$ and (iii) a number $0 \le r \le 1$ (the ratio of data from all classes) we constructed $n$ subsets of $\mathcal{D}$: $\left\lbrace\mathcal{D}_i\right\rbrace_{1 \le i \le n}$ to be used for optimisation during the $n$ pretraining stages. In doing this we applied the following methodology:
\begin{enumerate}
    \item We created a partition of the all classes: $\left\lbrace \mathcal{C}_1, \ldots \mathcal{C}_{n + 1} \right\rbrace$ such that $\mathcal{C}_i \cap \mathcal{C}_j = \emptyset, \forall i, j$, and $\mathcal{C} = \bigcup_{i=1}^{n+1} \mathcal{C}_{i}$.
    \item We randomly split the full training set $\mathcal{D}$ in two: $\mathcal{D}_{c}$, $\mathcal{D}_{u}$ such that $\frac{\vert\mathcal{D}_{u}\vert}{\vert\mathcal{D}\vert} = r$. (of course: $\mathcal{D}_{u} \cap \mathcal{D}_{c} = \emptyset$, and $\mathcal{D}_{u} \cup \mathcal{D}_{c} = \mathcal{D}$).
    \item We partition $\mathcal{D}_{u}$ into $n+1$ subsets: $\left\lbrace D_{u,1}, \ldots \mathcal{D}_{u, n+1} \right\rbrace$.
    \item We now define the data sets used to optimise the model in each stage (and considering $\mathcal{D}_{0}=\emptyset$):
    $$ \mathcal{D}_i = \mathcal{D}_{i-1} \cup \mathcal{D}_{u, i} \cup \left\lbrace \left(x, c\right) \in \mathcal{D}_{c} \mid c \in \mathcal{C}_i \right\rbrace$$
\end{enumerate}

The $n+1$-th dataset $\mathcal{D}_{n+1} \equiv \mathcal{D}$ corresponds to the final tuning phase on the full data set.

\subsection{Residual networks of various depths and widths}
\label{sec:resnets}

In the experiment with residual neural networks of different widths and depths we changed the architecture of ResNet-18 \cite{He_2016_CVPR} as follows.

Apart from the first convolution and the fully connected layer at the output, ResNet-18 consists of four modules, each made up of $d=2$ residual blocks with the same number of output channels. Each module doubles the number of channels and halves the height and the width of the feature maps. The first module receives a volume with $w=64$ channels, the second operates on $2w=128$, and so on.

In our experiments we uniformly changed the depth $d$ of the four modules, and/or scaled the number of channels in all modules ($w, 2w, 4w, 8w$).

Note that this is not the standard way in which people design deeper residual architectures such as ResNet-32, or ResNet-55. Deeper ResNets increase the depth of the modules differently, and use bottleneck blocks to avoid an explosion in the number of parameters.

\subsection{Resetting the layers of the model}
\label{sec:reset_details}

In the experiments presented in Figure~\ref{fig:reset} from Section~\ref{sec:results} we reset subsets of the model's parameters. We reset entire modules referring with 1 to the first convolution, with numbers from 2 to 5 to the four modules (each consisting of 2 residual blocks), and naming 6 the last fully connected layer.

\begin{figure}[h!tb]
    \centering%
    \includegraphics[width=\linewidth]{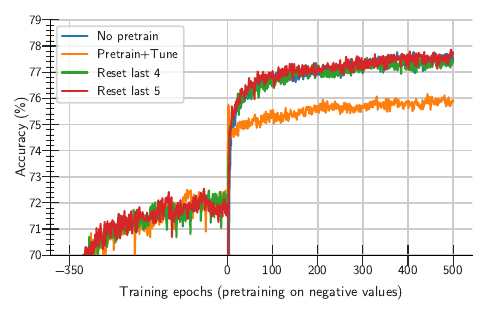}
    \caption{Here we show the learning curves for three models pretrained for 350 epochs on half of data. For two of them we keep the first 1 or 2 modules, reinitialise the rest and tune for 500 epochs.}
    \label{fig:advantage350}
\end{figure}

As Figure~\ref{fig:reset} shows, resetting the last 4, or 5 modules seems to recover the original performance of a model trained from random parameters. Therefore we asked whether keeping the pretrained parameters of the first 1 or 2 modules comes with any advantage in terms of training speed. As Figure~\ref{fig:advantage350} shows, in our setup there seems to be no benefit from preserving parameters from the pretraining stage.

\section{Supporting evidence for the two phases of learning hyothesis}
\label{sec:twophase_literature}

A couple of works identify critical differences between the early stage and the late stage of training, offering empirical evidence for the two phases of learning hypothesis.

\cite{achille2018critical} identifies an initial \textit{memorisation} phase when data information is absorbed into the network's weights, followed by a \textit{reorganisation} stage where unimportant connections are pruned and information decreases while being redistributed among layers for efficiency. \citeauthor{achille2018critical} used the Fisher Information Matrix to approximate the amount of information stored in the weights. The FIM is also a curvature matrix, therefore the observed regimes support the view that learning changes basins of attraction of different minima until it lands in one with low curvature, corresponding to a flat minimum. \citeauthor{achille2018critical} also point out that if data statistics change after the initial phase, the network would remain trapped in the valley the memorisation phase guided it into.

\cite{golatkar2019time} empirically shows that regularisation has an impact on final generalisation performance only in the early stages of training. Applying weight decay or data augmentation only after this initial phase, or stopping regularisation after that point would not affect generalisation. The experiments using data augmentation later in training offer additional evidence for the generalisation gap -- if one thinks about that stage as tuning on more data from the same distribution.

\cite{gur2018gradient} shows that after an early training stage the gradients reside in a small subspace that remains constant for the rest of training. This reiterates the importance of the data used in the first steps of training.

\cite{li2019towards} shows that in overparametrized networks the volume of good minima dominates the volume of poor minima and underlines the importance of a high learning rate to land in the basin of attraction of a well generalising minimum of the loss function. \cite{jastrzebski2020break} extends the observation with the importance of using batches to induce the noise needed to escape poorly generalising minima. Precisely, \citeauthor{jastrzebski2020break} point out that the ratio between learning rate and batch size determines the flatness of the minumum.

\cite{ghorbani2019investigation} computes the full spectrum of the Hessian showing that \citeauthor{jastrzebski2020break}'s claims about smaller learning rates guiding the network into sharper minima doesn't hold empirically. A possible explanation is that the network is already trapped around some minima, and the slow learning rate just reaches an even flatter region closer to the critical point.

\end{document}